\documentclass[letterpaper, 10 pt, conference,table]{ieeeconf}  

\IEEEoverridecommandlockouts                              

\usepackage{graphicx}
\usepackage{lipsum}
\usepackage[linesnumbered,ruled,vlined]{algorithm2e}
\usepackage{amssymb,amsmath,latexsym,amsfonts}
\usepackage[usestackEOL]{stackengine}
\usepackage{amssymb}
\usepackage{caption,url}
\usepackage{float}
\usepackage{epstopdf}
\usepackage{authblk}
\usepackage{dblfloatfix}
\usepackage{multirow}
\usepackage{wrapfig}
\usepackage{cite}

\usepackage{hyperref}
\hypersetup{colorlinks,breaklinks,urlcolor=blue}
\usepackage{dblfloatfix}
\usepackage{multirow}

\usepackage[
  separate-uncertainty = true,
  multi-part-units = repeat
]{siunitx}

\usepackage{subcaption}
\usepackage{gensymb}
\usepackage{tikz}
\usepackage{hhline}

\definecolor{darkgreen}{rgb}{0.0, 0.2, 0.13}

\title{\LARGE \bf
  Memory based neural networks for end-to-end autonomous driving
}

\author{{Sergio Paniego Blanco, Sakshay Mahna, Utkarsh A. Mishra,  Jos\'eMar\'ia Ca\~nas}

\thanks{Address all correspondance to: \href{mailto:sergio.paniego@urjc.es}{sergio.paniego@urjc.es}}
\thanks{The presented work is partially supported by Google Summer of Code initiative.}
\thanks{S. P. Blanco and J. Ca\~nas are with Universidad Rey Juan Carlos, Madrid, Spain and JDErobot organization.}
\thanks{S. Mahna and U. A. Mishra are student members of the JDErobot organization.}
}

\begin{document}

\maketitle
\thispagestyle{empty}
\pagestyle{empty}

\begin{abstract}


Recent works in end-to-end control for autonomous driving have investigated the use of vision-based exteroceptive perception. Inspired by such results, we propose a new end-to-end memory-based neural architecture for robot steering and throttle control. We describe and compare this architecture with previous approaches using fundamental error metrics (MAE, MSE) and several external metrics based on their performance on simulated test circuits. The presented work demonstrates the advantages of using internal memory for better generalization capabilities of the model and allowing it to drive in a broader amount of circuits/situations. We analyze the algorithm in a wide range of environments and conclude that the proposed pipeline is robust to varying camera configurations. All the present work, including datasets, network models architectures, weights, simulator, and comparison software, is open source and easy to replicate and extend. Code: \href{https://www.github.com/JdeRobot/DeepLearningStudio}{github.com/JdeRobot/DeepLearningStudio}.

\end{abstract}

\section{INTRODUCTION}

Making cars and robots that are capable of driving by itself has been a topic of interest on both research and industry for the past years and seems to be a topic that will have a wide impact on day-to-day life in the coming years too \cite{Litman2022}. The potential benefits of autonomous driving and robots is enormous, including improved traffic safety and security or a more optimized mobility for individuals and globally.

Typically, this autonomy is divided into 5 levels, from no automation (0) to fully automation (5), where human intervention is no needed in any situation. Some current commercial solutions (e.g. Tesla, Waymo, etc.) implement level 2 and even 3 autonomy capabilities, but most benefits come on level 4 and 5, which still need advancements in research and industry \cite{Litman2022}. Some competitions are also helping advance this research, including DuckieTown\cite{Paull2017} and AWS DeepRacer\cite{Balaji2019}.

This interest has experimented a rise thanks to the last advancements in deep learning, specially with the inclusion of specialized hardware (GPUs), the development of large open datasets \cite{Cabon2020} and the advancement in previous techniques, such as CNN \cite{LeCun1989}. Deep learning and the solutions based on artificial intelligence help improve the results on this perception and control problem. 

\begin{figure}[t]
\includegraphics[width=0.9\linewidth]{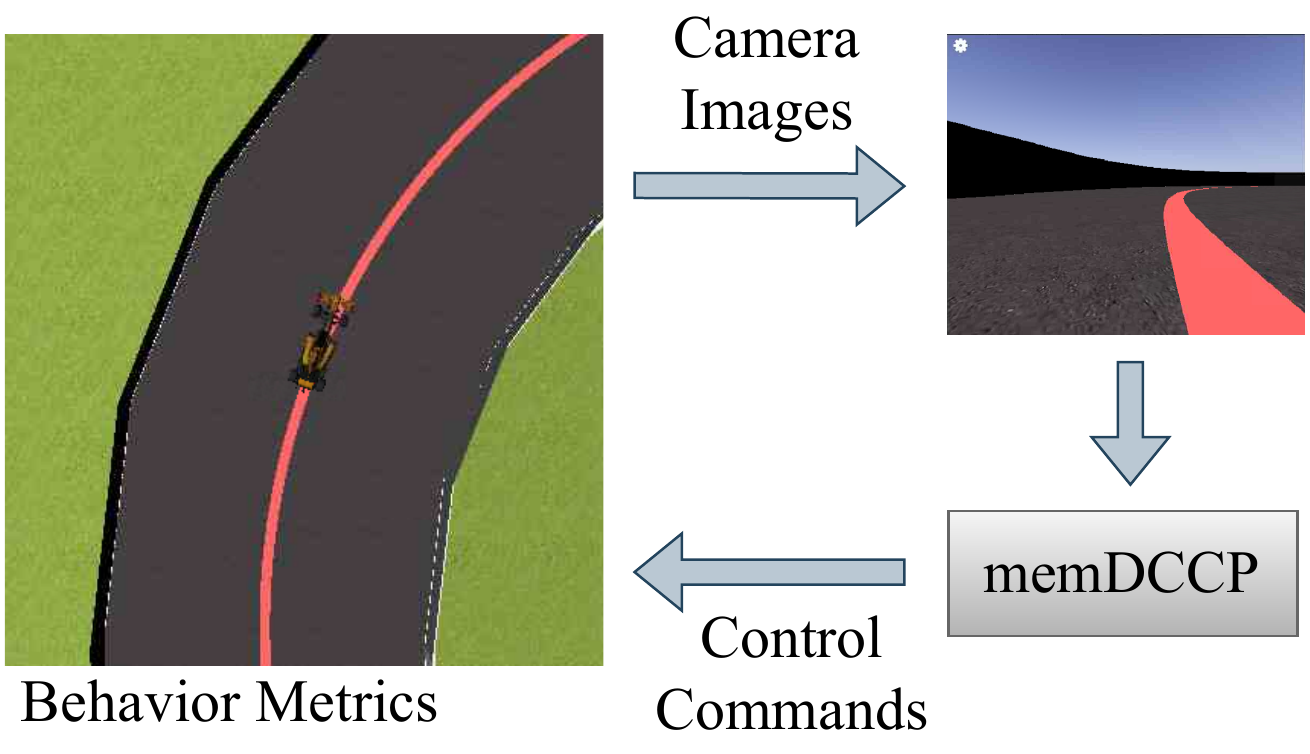}
\centering
\caption{End-to-end autonomous driving pipeline using Behavior Metrics framework and the proposed DeepestConvLSTMConv3DPilotnet (memDCCP) architecture.}\label{intro}
\end{figure}

The methodologies for autonomous driving control solutions are usually divided between modular and end-to-end pipelines. The first one includes several modules whereas the end-to-end \cite{Bojarski2016, Bojarski2017} pipeline generates decisions based on the direct understanding of the input (typically images). Vision based end-to-end autonomous driving have gained significant attraction in recent literature. Vision is an integral segment of accumulating exteroceptive information about the surrounding environment of the vehicle~\cite{Riboni2021, Zhao2020, Pomerleau1988, Bojarski2017, Xu2016, Santana2016, Liu2020, Kocic2019, Ly2021}. Using vision for end-to-end control of autonomous vehicles was proposed by NVIDIA's PilotNet~\cite{Bojarski2017} framework where the goal was to deliver steering and control commands based on raw frontal images with the help of a CNN. This was further analyzed by augmenting fully connected layers to a CNN~\cite{Rausch2017} to achieve performance comparable to a human driver.  Simulated images have also been used by~\cite{yang2017feature} to study the importance of road related features in the images. While all the above contributions considered utilizing the spatial features of the camera image inputs, \cite{Chi2017} proposed addition of temporal analysis using memory based DNNs. The proposed methods, namely LSTM and Conv-LSTM PilotNet, examined the importance of LSTMs and convolutional layers with LSTMs respectively.

The use of memory based solutions has also been explored previously on different research publications \cite{delEgio2019} \cite{Eraqi2017}. Long Short-Term Memory networks (LSTMs) \cite{Hochreiter1997} are a special type of recurrent neural networks (RNN), that are specialized in learning temporal dependencies that are present in the dataset. A variation of the LSTMs are convolutional LSTM (ConvLSTMs) \cite{Shi2015}, architectures that include convolutions and are able to learn these temporal dependencies from sequences of images. 3D convolutions (CONV3Ds) are another commonly used architectural layer when the temporal dependencies are important to be learned. 

When understanding the performance of different solutions for end-to-end control of an autonomous car/robot, loss metrics are very important. Some commonly used metrics include mean average error (MAE) and mean square error (MSE). These metrics show the performance of the models on a test set of the dataset. In addition to them, the actual performance of the model on test circuits/environments can also be interesting for a better knowledge of the system but difficult to quantify. When having closed circuits, these external metrics could include whether the model is able to complete a lap on a test circuit or the time per lap.

In this paper, we present and describe a deep learning architecture based on ConvLSTM and Conv3D, DeepestConvLSTMConv3DPilotnet (memDCCP), that is capable of driving autonomously over different circuits and its variations, improving the performance and generalization of networks that are memory-less and approaches that are not based on deep learning. In addition, we present a robustness analysis of the different networks, to understand their strengths. We describe a comparison using internal metrics (MAE, MSE) but we also describe a comparison using external metrics obtained from an open source application, also developed by us, Behavior Metrics (Figure \ref{intro}). These external metrics are complementary to the internal ones and describe how well do the models really perform in simulations on circuits that they have not seen on the dataset.





\section{ROBOT CONTROL BASED ON NEURAL NETWORKS}

This section is devoted to explain the system we have created, with different deep learning based approaches that we have used for training a car on driving autonomously over circuits, learning from a dataset. The simulator structure and its corresponding application for extracting the external metrics is explained in the following section.

There is a division on the deep learning models used in the present work, memory-less models and memory based models. The first group of models use an instant perception in order to generate the predicted linear and angular speeds. For each image that the car perceives, the model outputs both values. The second group includes models that use sequences of images to generate the predictions. The temporal factor is used in this case, the models receive more than one image as input, the current image and some previous frames (3 frames in the case of our experiments), and with this information it predicts linear and angular speeds, following a many-to-one approach. 

In order to generate the dataset for training and evaluation, we have generated a explicitly programmed brain based on a PID controller that is able to drive on different circuits. These circuits have a red line on the middle of the road along the whole extension of the circuit that serves as a guide. The explicitly programmed brain has a filter that looks for that red line in the images that it captures and based on the position of the red line, it generates linear and angular speeds. A brain is the program responsible for generating angular and linear speeds for the car based on the sensor information, camera images in this case.

The dataset consist of images gathered by this explicitly programmed brain, running over four different F1 circuits. For testing, the circuits used are variations of some of them (circuit without red line, circuit with different road color...) and F1 circuits not used for the dataset. 

\subsection{Memory-less deep learning models}

In this group of models, PilotNet \cite{Bojarski2016} and DeepestLSTMTinyPilotNet \cite{delEgio2019} are tested. Both models have been described in previous works on end-to-end learning of throttling for self-driving cars. They receive one image at a time as input, run the inference on the image, applying cropping to remove the upper part of the image, and generate predictions for angular speeds. The architectures used in this work are the ones described in these previous works, with the small modification on the output neuron that instead of having one neuron just for angular speed, the models output two values, for the linear and angular speed.

\begin{itemize}
  \item \textbf{PilotNet}: powerful network with a series of convolutional layers followed by some fully connected layers that finally end on two output neurons.
  \item \textbf{DeepestLSTMTinyPilotNet}: modification of the PilotNet model making it smaller, reducing the number of convolutional and fully connected layers. It has some ConvLSTM layers that add some memory information, but since it only used one image as input, it still remains in the memory-less group.
\end{itemize}

\subsection{Memory-based deep learning models}

Two models are presented and tested as memory-based options. They are based on the memory-less models, just adding small modifications to make them able to process multiple frames as the same time (sequences of images). All of them receive sequences of 3 images temporally dependent and predict two values, linear and angular speeds, with a many-to-one approach. Both of them have been developed for the project, although only one of them has resulted in good results, memDCCP.

\begin{figure}[H]
\includegraphics[width=0.65\linewidth]{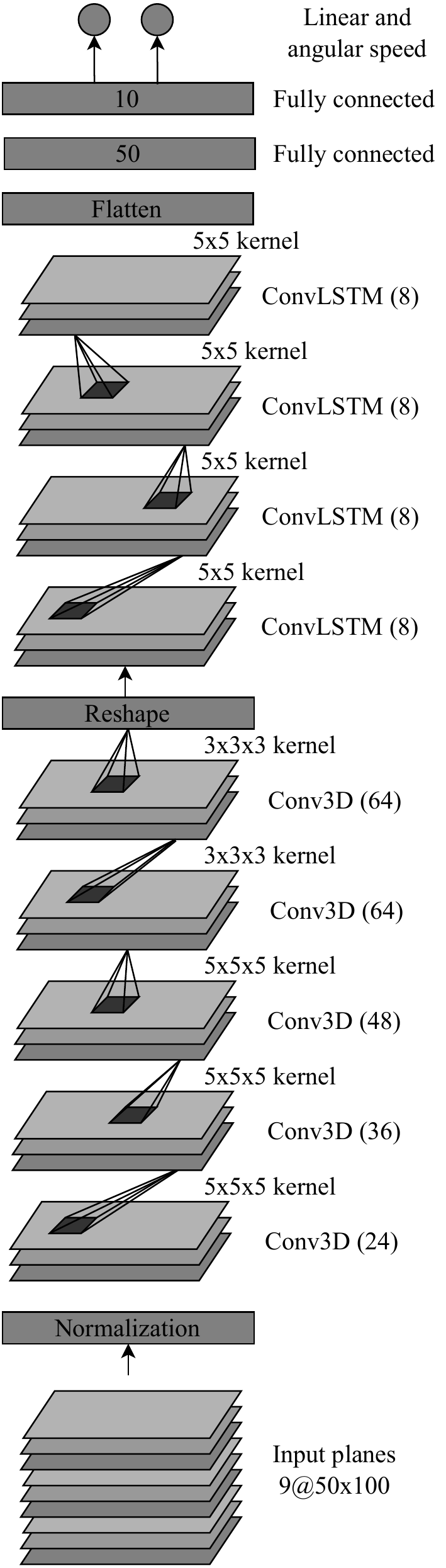}
\centering
\caption{Proposed DeepestConvLSTMConv3DPilotnet (memDCCP) architecture combining Conv3D and ConvLSTM layers to extract temporal information and help in the prediction of control commands.}\label{memDCCP}
\end{figure}

\begin{itemize}
  \item \textbf{PilotNet x3}: the architecture is exactly the same as in PilotNet, but it receives sequences of 3 images instead of image each time having a layer wrapper on the convolutional layers that allows performing convolutions to the 3 images of each sequence.
  \item \textbf{memDCCP}: evolution based on DeepestLSTMTinyPilotNet ideas. On the first layers of the network, 5 Conv3D layers are used, followed by 3 ConvLSTM layers and finishing with 3 fully connected layers on the end of the architecture. In Figure \ref{memDCCP}, a detail of the architecture is illustrated. This network architecture is the one that was found to outperform the rest of the models.
\end{itemize}

\section{MEASURING NEURAL NETWORKS FOR ROBOT CONTROL}

A complete application has been developed for measuring the performance of the different brains, Behavior Metrics, that connects to the simulator, Gazebo, and extracts the metrics results using ROS.

When measuring the performance of solutions for robot control, common metrics like MAE and MSE are commonly used. In the present work, these metrics receive the name of internal metrics. The information that they provide can be complemented with external metrics, which are results from actually running simulation experiments using brains on test environments/circuits. These metrics are generated and analyzed internally by Behavior Metrics.

For the problem of end-to-end control in autonomous driving, previous decisions play an important role on the current scenario. For example, if the car predicted a wrong decision some previous steps in time ago, the situation at the current time can be very different from the one where the car has driven smoothly over the whole circuit. Another example would be when the car drives really fast and a sharp curve is some steps in time ahead. The linear and angular speed will be playing an important role when the curve is approaching and the car has to modify its behavior, slow down and manage to overcome the curve. 

These types of situations are difficult to study solely measuring MAE and MSE over the test dataset. To help understanding the behavior of the control solutions, we have developed Behavior Metrics~\cite{behaviorMetrics} based on the Gazebo simulator. The application serves as the graphical interface for running simulations, switching between circuits, saving and analyzing metrics, and visualizing camera readings.

\begin{figure}[h]
\includegraphics[width=\linewidth]{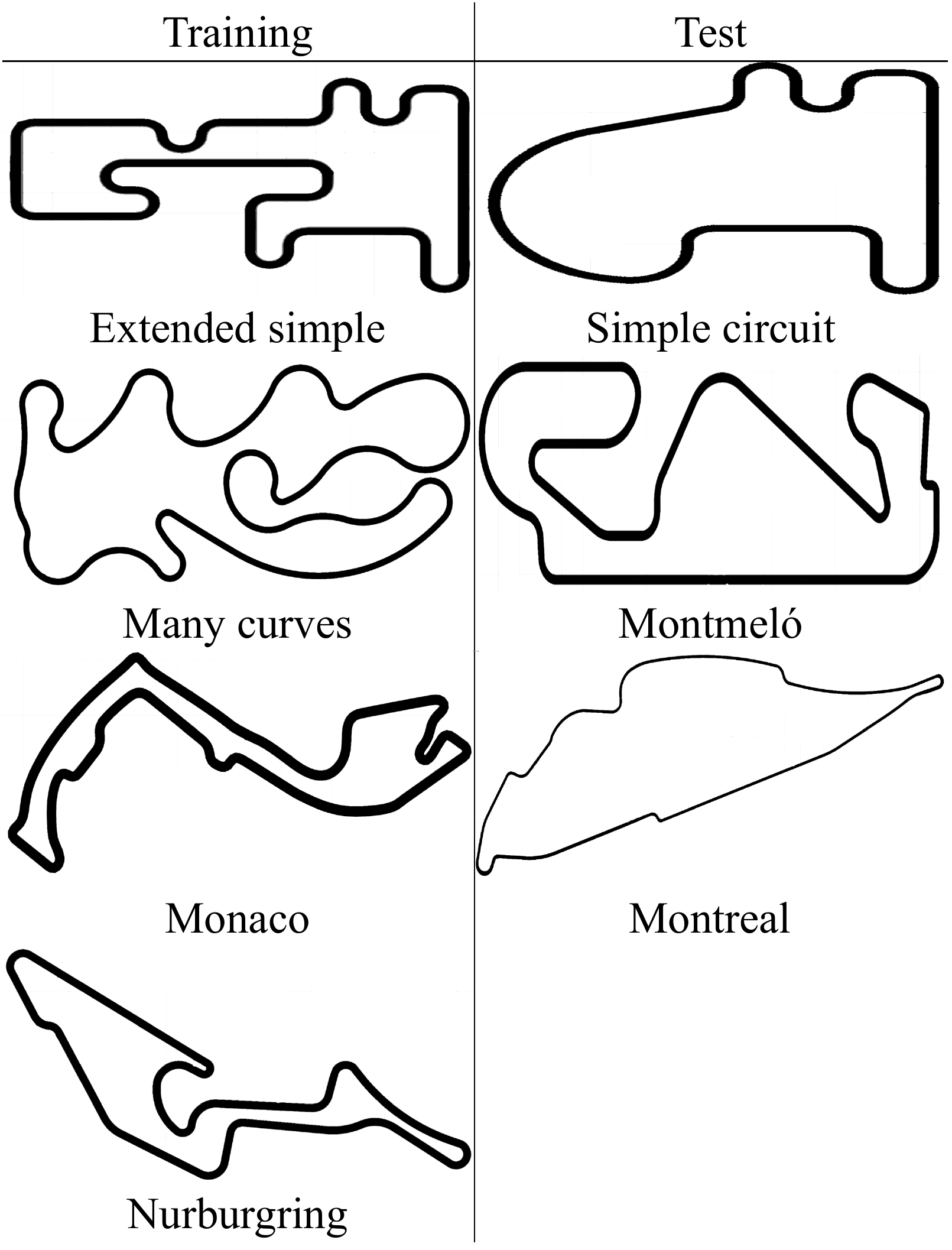}
\centering
\caption{We incorporate seven significantly different circuits for our analysis and present their respective silhouettes.}\label{all_circuit}
\end{figure}

Behavior Metrics includes several F1 circuits for simulation, as illustrated in Figure \ref{all_circuit}. The default version of all of them has a red line on the middle that traverses the whole circuit and is used by the explicitly programmed brain to drive and generate the dataset from which the neural brains learn. In addition, each circuit has the possible following variations:

\begin{itemize}
  \item Line color: red, white or no line.
  \item Road texture: asphalt (grey) or white.
  \item Lateral walls: present or not present.
\end{itemize}

When simulating, Behavior Metrics measures a set of external metrics for each experiment that complement the internal metrics. These metrics include average speed, time per lap and position deviation MAE. The position deviation metric indicates how well does the car follows the red line (if included on the circuit) calculating the distance from the position of the car to the line each time.

Having a batch of circuits with different variations and shapes can help understanding the generalization capabilities that the models learn. These circuits have different types of straight and curved parts with a range of difficulty levels and come with different road textures and lighting conditions. Some of them are inspired in real world F1 circuits, for example Montmeló or Monaco. Since they are based on real world F1 circuits, they provide a broad set of situations, for example extremely difficult curves or sharp curves after long straights parts.

The dataset used for training has been generated using the explicitly programmed brain, driving following the red line on the middle of the circuits over 4 of the circuits that Behavior Metrics provides (detailed in Figure \ref{all_circuit}), leaving 3 circuits for testing.

\section{EXPERIMENTS}

This section is dedicated to analyze and compare the performance of the different presented end-to-end brains. Three experiments have been conducted, including the comparison of neural network models with the explicitly programmed brain on different scenarios, a generalization analysis describing how well the neural brains are able to drive in a variety of situations compared to the explicitly programmed brain and finally, a robustness analysis. 

All the  different experiments are easily reproducible, with the models weights, architectures and the  application for simulation and experiment available open-source. The structure of the simulator and networks has been previously presented, featuring some commonly used software tools like Gazebo and ROS on the simulation part. Tensorflow-Keras has been used for programming and training the different deep learning architectures.

\subsection{Comparison with explicitly programmed brain}

In the first experiment, we compare and analyze the performance of the explicitly programmed brain versus the neural network controlled brains in different circuits that have not been used for generating the dataset used for training. 

In Table~\ref{mae_mse_table}, a comparison of internal metrics (MAE, MSE) after training for 300 epochs is displayed. The dataset used for training all the networks, those with memory and those without memory is essentially the same. The same preproccessing steps have been followed for all the approaches (cropping the images on the horizon line and mirroring) and during training, some data augmentation techniques have been used. The only difference between the training of the memory-less networks and the memory-based ones is that for the second group, the images are fed as a sequence of three images each time, all of them sharing a temporal dependency (the current frame, the previous one and the frame of 2 time steps ago).

The values are essentially lower for PilotNet style networks for both metrics. This could mean that PilotNet style networks learn the dataset better and could behave better when tested. This fact can be additionally explained connecting with table \ref{param_number}, where the number of parameters for each model is displayed.

\begin{table}[h]
\caption{MAE and MSE metrics comparison}
\label{mae_mse_table}
\begin{center}
\begin{tabular}{|c|c|c|c|c|}
\hline
Model & MAE train & MAE val & MSE train & MSE val \\
\hline
PilotNet & 0.0044 & 0.00546 & \textbf{0.00016} & 0.00064 \\
\hline
\shortstack{Deepest  \\ LSTM \\ TinyPilotNet} & 0.018 & 0.016 & 0.00416 & 0.00382\\
\hline
\shortstack{PilotNet  \\ x3 } & \textbf{0.0044} & \textbf{0.0048} & 0.00018 & \textbf{0.00032} \\
\hline
\shortstack{memDCCP} & 0.019 & 0.018 & 0.0039 & 0.0035 \\
\hline
\end{tabular}
\end{center}
\end{table}

PilotNet style networks are the deep learning model that have the highest number of parameters to learn from the data, so that can explain why it is the network that achieves the best performance on the internal metrics. In the following experiment, where the generalization is studied, these results are shown not to be enough to achieve a good performance in certain scenarios. 

\begin{table}[h]
\caption{Number of parameters}
\label{param_number}
\begin{center}
\begin{tabular}{|c|c|c|}
\hline
Model & Number of parameters & Input shape \\
\hline
PilotNet & 17M & 200x66\\
\hline
\shortstack{Deepest  \\ LSTM \\ TinyPilotNet} & 62k & 100x50 \\
\hline
\shortstack{PilotNet  \\ x3 } & 15.9M & 100x50 \\
\hline
\shortstack{memDCCP} & 910k & 100x50 \\
\hline

\end{tabular}
\end{center}
\end{table}

In table \ref{comparison_explicit}, a comparison of performance between the explicitly programmed brain and the neural networks based brains is displayed. This comparison is done using a testing circuit, Montmeló. This circuit has some difficult curves, even including a chicane. Each experiment is conducted 3 times, averaging the results.

\begin{table}[h]
\caption{Comparison with explicitly programmed brain in Montmeló}
\label{comparison_explicit}
\begin{center}
\begin{tabular}{|c|c|c|c|}
\hline
Model & Lap seconds &  \shortstack{MAE position \\ deviation} & Average speed \\
\hline
\shortstack{Explicitly \\ programmed \\ brain}  & 78s & 3.01 & 9.34m/s \\
\hline
PilotNet & \textbf{77s} & \textbf{2.73} & \textbf{9.43m/s}\\
\hline
\shortstack{Deepest  \\ LSTM \\ TinyPilotNet} & \textbf{77s} & 4.76 & 9.17m/s \\
\hline
\shortstack{PilotNet  \\ x3 } & 97.3s & 5.40 & 7.52m/s\\
\hline
\shortstack{memDCCP} & 80s & 4.93 & 9.17m/s \\
\hline
\end{tabular}
\end{center}
\end{table}






The comparison shows slightly better results for PilotNet model on the time spent per lap, position deviation MAE and average speed over the experiments, but the numbers are really close to the ones achieved by the explicitly programmed brain. This can be explained because the dataset generated for training has been generated using the explicitly programmed brain, so the actions that the neural brains learn are close to the ones generated by the explicitly programmed brain but they will hardly outperform its performance by a large number, at least in circuits that are close to the dataset considering its appearance. 

Curiously, the worst performance is achieved by PilotNet x3, in contrast to its results on internal metrics. This network is slower and follows the red line worse than the other ones.

If we consider circuits whose appearance is further from the dataset and from the explicit actions programmed for the explicitly programmed brain, the results are different, as the following experiment covers.

\subsection{Generalization analysis}
 
 For the second experiment, we tested the generalization of the different solutions for end-to-end control of the linear and angular speeds of the robot. Generalization refers to how the solutions are able to drive on a broader set of situations that are new for the system. 
 
 Basically, we tested if the robot was able to complete laps on variations of circuits using the explicitly programmed brain and the neural network based brains. In table \ref{generalization_table}, a relation of brains and time per lap using the simple circuit is shown.
 
 \begin{table}[h]
\caption{Generalization comparison in simple circuit}
\label{generalization_table}
\begin{center}
\begin{tabular}{|c|c|c|c|c|c|c|}
\hline
\textbf{Config.} & \multicolumn{6}{c|}{} \\
\hline
\shortstack{Line}  & red & red & white & no   & no  &  no\\
\hline
\shortstack{Road \\ color}  & grey & white & grey & grey  & white   & grey\\
\hline
\shortstack{Lateral \\ walls}  & yes & yes & yes  & yes & yes  & no\\
\hline
\textbf{Brain} & \multicolumn{6}{c|}{} \\
\hline
\shortstack{Explicitly  \\ programmed \\ brain}  & \textcolor{darkgreen}{50s}\textbf{\textcolor{darkgreen}{61s}}
 & \textbf{\textcolor{darkgreen}{50s}} & {\color{red} -} & {\color{red} -} & {\color{red} -}& {\color{red} -}\\
\hline
PilotNet & \textcolor{darkgreen}{50s} & \textcolor{darkgreen}{55s} & \textbf{\textcolor{darkgreen}{50s}} & {\color{red} -} & {\color{red} -} & {\color{red} -} \\
\hline
\shortstack{Deepest  \\ LSTM \\ Tiny \\PilotNet} & \textbf{\textcolor{darkgreen}{46s}} & \textcolor{darkgreen}{53s} & {\color{red} -} & {\color{red} -} & {\color{red} -}  & {\color{red} -} \\
\hline
\shortstack{PilotNet  \\ x3 } & \textcolor{darkgreen}{54.5s} & \textcolor{darkgreen}{60s} & {\color{red} -} & {\color{red} -} & {\color{red} -} & {\color{red} -} \\
\hline
\shortstack{memDCCP} &  \textcolor{darkgreen}{56s} & \textcolor{darkgreen}{61s} & \textcolor{darkgreen}{62.6s} & \textbf{\textcolor{darkgreen}{72s}} & \textbf{\textcolor{darkgreen}{61s}} & {\color{red} -} \\
\hline
\end{tabular}
\end{center}
\end{table}

 The circuits used in this experiment are variations of the simple circuit. In addition to the circuit with the red line that crosses it on the middle of the road along the whole path, essential for the explicitly programmed brain to drive, we introduce variations. These variations include removing or changing the color of the red line, removing the walls on the side of the circuit (in Figure \ref{intro} these walls are present, so when the circuits does not have them, the robot will see the grass on the sides of the circuit) or changing the road color to a white texture. All the variations are present on the table. Since the car bases its behavior on the understanding of the image it captures from the road, these variations are interesting for testing the generalization.
 
 Looking at the table, we can clearly see that the brain that generalizes better and that is able to complete more circuit variations is memDCCP, which is able to complete almost all the variations. On the other side, the rest of the brain models, including the explicitly programmed one, can drive on the circuits that have a red line along the middle of the road, but they fail to drive on most of the variations when the line color is modified or is missing.

\subsection{Robustness analysis}

The robustness to noise and certain camera variations was studied in the third experiment. The different candidate brains were tested on test circuits (circuits not used for training) to see how well they could cope up with random fluctuations with the robot hardware.

Table \ref{offset_camera_table} illustrates the performance of the different brains on slightly changing the camera positions on the robot. Table \ref{noise_table} shows the performance of the different brains when different levels of salt and pepper noise is present in the image received from the robot camera.

All the brains up to some level, were able to cope with the different fluctuations. The memDCCP, was able to deal with both the camera offsets and the noise.

\begin{table}[h]
\begin{center}
\caption{Comparison of Brains with Camera Offset}
\label{offset_camera_table}
\resizebox{\columnwidth}{!}{%
\begin{tabular}{|c|cc|cc|cc|}
\hline
\textbf{Offset} &
  \multicolumn{2}{c|}{\textbf{\begin{tabular}[c]{@{}c@{}}Camera \\ Moved \\ Left\end{tabular}}} &
  \multicolumn{2}{c|}{\textbf{\begin{tabular}[c]{@{}c@{}}Camera \\ Moved \\ Right\end{tabular}}} &
  \multicolumn{2}{c|}{\textbf{\begin{tabular}[c]{@{}c@{}}Camera \\ Rotated \\ Down\end{tabular}}} \\ \hline
\textbf{Model} &
  \multicolumn{1}{c|}{\textbf{\begin{tabular}[c]{@{}c@{}}Lap \\ Seconds\end{tabular}}} &
  \textbf{MAE} &
  \multicolumn{1}{c|}{\textbf{\begin{tabular}[c]{@{}c@{}}Lap \\ Seconds\end{tabular}}} &
  \textbf{MAE} &
  \multicolumn{1}{c|}{\textbf{\begin{tabular}[c]{@{}c@{}}Lap \\ Seconds\end{tabular}}} &
  \textbf{MAE} \\ \hline
\begin{tabular}[c]{@{}c@{}}Explicit \\ Brain\end{tabular} &
  \multicolumn{1}{c|}{123} &
  5.227 &
  \multicolumn{1}{c|}{115} &
  17.155 &
  \multicolumn{1}{c|}{-} &
  - \\ \hline
Pilotnet &
  \multicolumn{1}{c|}{140} &
  4.715 &
  \multicolumn{1}{c|}{142} &
  20.99 &
  \multicolumn{1}{c|}{-} &
  - \\ \hline
\begin{tabular}[c]{@{}c@{}}Deepest\\ LSTM\\ TinyPilotnet\end{tabular} &
  \multicolumn{1}{c|}{160} &
  4.15 &
  \multicolumn{1}{c|}{159} &
  25.369 &
  \multicolumn{1}{c|}{160} &
  9.13 \\ \hline
\begin{tabular}[c]{@{}c@{}}Pilotnet\\ x3\end{tabular} &
  \multicolumn{1}{c|}{145} &
  4.548 &
  \multicolumn{1}{c|}{144} &
  24.787 &
  \multicolumn{1}{c|}{-} &
  - \\ \hline
\begin{tabular}[c]{@{}c@{}}memDCCP\end{tabular} &
  \multicolumn{1}{c|}{160} &
  6.418 &
  \multicolumn{1}{c|}{160} &
  28.792 &
  \multicolumn{1}{c|}{162} &
  10.363 \\ \hline
\end{tabular}%
}
\end{center}
\end{table}

\begin{table}[h]
\begin{center}
\caption{Comparison of Brains under Salt \& Pepper Noise }
\label{noise_table}
\resizebox{\columnwidth}{!}{%
\begin{tabular}{|c|cc|cc|cc|}
\hline
\textbf{Noise} &
  \multicolumn{2}{c|}{\textbf{\begin{tabular}[c]{@{}c@{}}Noise \\ Probability \\ 0.2\end{tabular}}} &
  \multicolumn{2}{c|}{\textbf{\begin{tabular}[c]{@{}c@{}}Noise \\ Probability \\ 0.4\end{tabular}}} &
  \multicolumn{2}{c|}{\textbf{\begin{tabular}[c]{@{}c@{}}Noise \\ Probability \\ 0.6\end{tabular}}} \\ \hline
\textbf{Brains} &
  \multicolumn{1}{c|}{\textbf{\begin{tabular}[c]{@{}c@{}}Lap \\ Seconds\end{tabular}}} &
  \textbf{MAE} &
  \multicolumn{1}{c|}{\textbf{\begin{tabular}[c]{@{}c@{}}Lap \\ Seconds\end{tabular}}} &
  \textbf{MAE} &
  \multicolumn{1}{c|}{\textbf{\begin{tabular}[c]{@{}c@{}}Lap \\ Seconds\end{tabular}}} &
  \textbf{MAE} \\ \hline
\begin{tabular}[c]{@{}c@{}}Explicit \\ Brain\end{tabular} &
  \multicolumn{1}{c|}{117} &
  5.474 &
  \multicolumn{1}{c|}{132} &
  5.401 &
  \multicolumn{1}{c|}{-} &
  - \\ \hline
Pilotnet &
  \multicolumn{1}{c|}{-} &
  - &
  \multicolumn{1}{c|}{-} &
  - &
  \multicolumn{1}{c|}{-} &
  - \\ \hline
\begin{tabular}[c]{@{}c@{}}Deepest \\ LSTM \\ TinyPilotnet\end{tabular} &
  \multicolumn{1}{c|}{-} &
  - &
  \multicolumn{1}{c|}{-} &
  - &
  \multicolumn{1}{c|}{-} &
  - \\ \hline
\begin{tabular}[c]{@{}c@{}}PilotNet \\ x3\end{tabular} &
  \multicolumn{1}{c|}{-} &
  - &
  \multicolumn{1}{c|}{-} &
  - &
  \multicolumn{1}{c|}{-} &
  - \\ \hline
\begin{tabular}[c]{@{}c@{}}memDCCP\end{tabular} &
  \multicolumn{1}{c|}{155} &
  9.958 &
  \multicolumn{1}{c|}{158} &
  8.52 &
  \multicolumn{1}{c|}{-} &
  - \\ \hline
\end{tabular}%
}
\end{center}
\end{table}

\section{CONCLUSIONS}

We have presented two deep learning models for end-to-end linear and angular speed control based on memory and proved that one of them, memDCCP, outperforms the rest thanks to its memory and combination of ConvLSTM and Conv3D layers. When using a memory based network, even having a small memory of three frames, the network is able to generalize and complete a broader set of circuits and its variations. We have also introduced a new way of comparing end-to-end solutions for autonomous driving using simulations on never seen circuits using Behavior Metrics, illustrating the importance of the external metrics as complementary information to the internal ones. Finally, we provide the whole project material, including simulator, experiments software (Behavior Metrics), network architecture code, weights and dataset as open source, for its easy reproducibility and extension.

\bibliographystyle{plain} 
\bibliography{refs} 

\end{document}